\RequirePackage{fix-cm}
\documentclass[twocolumn]{svjour3}          % twocolumn
\smartqed  % flush right qed marks, e.g. at end of proof
\usepackage{graphicx}
\usepackage{amsmath}
\usepackage{blindtext}
\usepackage{amsfonts}
\usepackage{amssymb}
\usepackage{xcolor}
\newcommand{\vu}{\mathbf{u}}
\newcommand{\vc}{\mathbf{c}}
\newcommand{\vl}{\mathbf{l}}
\newcommand{\vt}{\mathbf{t}}
\newcommand{\vp}{\mathbf{p}}

\begin{document}

\title{Optimal Triangulation Method is Not Really Optimal
%\thanks{Grants or other notes
%about the article that should go on the front page should be
%placed here. General acknowledgments should be placed at the end of the article.}
}

%\subtitle{Do you have a subtitle?\\ If so, write it here}

%\titlerunning{Short form of title}        % if too long for running head

\author{Seyed-Mahdi Nasiri\and
        Reshad Hosseini\and
        Hadi Moradi %etc.
}

%\address[1]{School of ECE, College of Engineering, University of Tehran, Tehran, Iran}
%\address[2]{School of Computer Science, Institute of Research in Fundamental Sciences (IPM), Tehran, Iran}
%\address[3]{Intelligent Systems Research Institute, SKKU, South Korea}

%\authorrunning{Short form of author list} % if too long for running head

\institute{SM. Nasiri \at
              \email{s.m.nasiri@gmail.com}\\           %  \\
              \emph{School of ECE, College of Engineering, University of Tehran, Tehran, Iran} \\
%             \emph{Present address:} of F. Author  %  if needed
           \and
           R. Hosseini \at
              Tel.: +98-21-82089799\\
              \email{reshad.hosseini@ut.ac.ir}\\
              \emph{School of ECE, College of Engineering, University of Tehran, Tehran, Iran}\\
              \emph{School of Computer Science, Institute of Research in Fundamental Sciences (IPM), Tehran, Iran}\\
           \and
           H. Moradi \at
              \email{hadi.moradi@ut.ac.ir}\\
              \emph{School of ECE, College of Engineering, University of Tehran, Tehran, Iran}\\
              \emph{Intelligent Systems Research Institute, SKKU, South Korea}
}

\date{Received: date / Accepted: date}
% The correct dates will be entered by the editor

%%***********************************************************
%% *** THIS puts a figure between title and abstract ***
%\twocolumn
%[{%
%    \renewcommand\twocolumn[1][]{#1}%
%    \maketitle
%    \begin{center}
%        \centering
%        \includegraphics[width=.85\textwidth]{imgs/titleTeX.jpg}
%        \vspace*{-3mm}
%        \captionof{figure}
%        {
%          The goal of this work is to localize a query photograph (left) by finding other images of the same place in a large geotagged image database (right).
%          We cast the problem as a classification task and learn a classifier for each location in the database.
%          We develop two procedures
%          to calibrate the outputs of the large number of per-location classifiers without the need for additional labelled training data.
%%          We develop a non-parametric \textcolor{myGrey}{(can we asy tis for FV w-norm norm?)} procedure to calibrate the outputs of the large number of per-location classifiers without the need for additional positive training data.
%        }
%    \end{center}%
%}]
%% ******************************************************

\maketitle

\begin{abstract}
Triangulation refers to the problem of finding a 3D point from its 2D projections on multiple camera images.
For solving this problem, it is the common practice to use so-called optimal triangulation method, which we call the $L_2$ method in this paper. But, the method can be optimal only if we assume no uncertainty in the camera parameters. Through extensive comparison on synthetic and real data, we observed that the $L_2$ method is actually not the best choice when there is uncertainty in the camera parameters. Interestingly, it can be observed that the simple mid-point method outperforms other methods. Apart from its high performance, the mid-point method has a simple closed formed solution for multiple camera images while the $L_2$ method is hard to be used for more than two camera images. Therefore, in contrast to the common practice, we argue that the simple mid-point method should be used in structure-from-motion applications where there is uncertainty in camera parameters.
\keywords{Triangulation \and Structure-from-Motion \and Mid-point method}
% \PACS{PACS code1 \and PACS code2 \and more}
% \subclass{MSC code1 \and MSC code2 \and more}
\end{abstract}

\section{Introduction}\label{intro}
\begin{figure*}
    \centering
    \includegraphics[width=0.9\textwidth,clip,keepaspectratio]{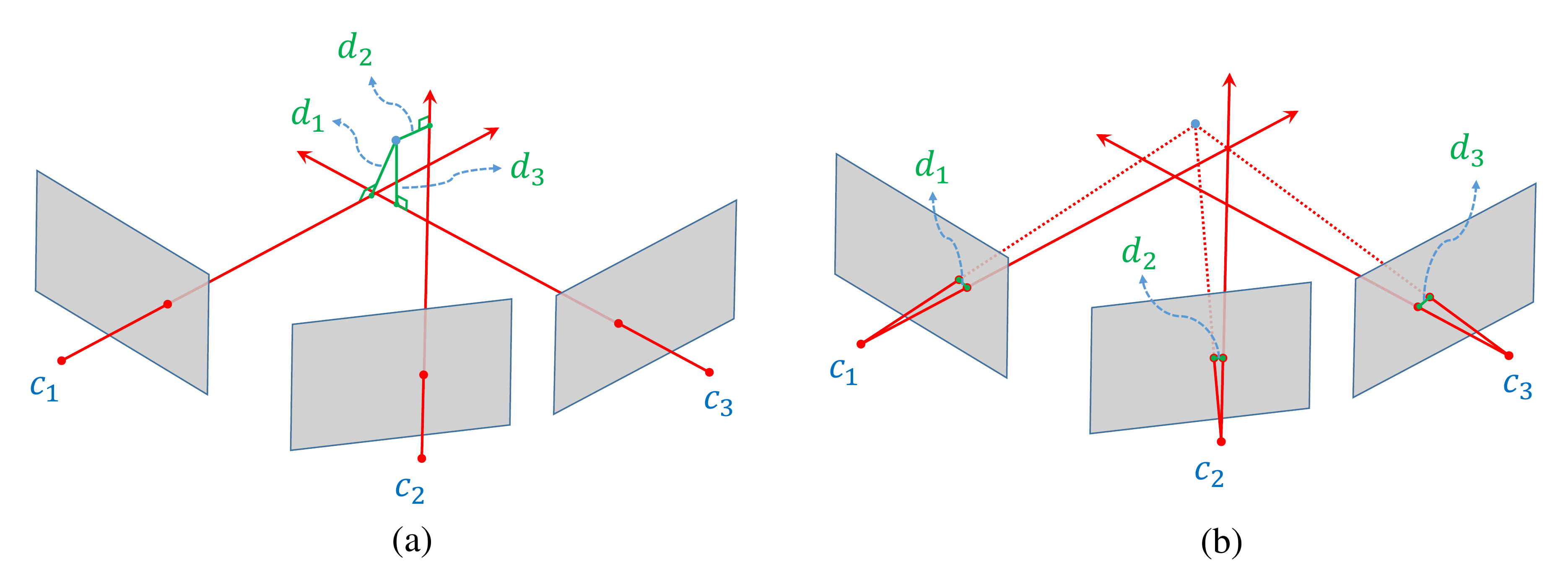}
    \caption{Triangulation methods: a)~Mid-point method finds a 3D point which has the minimum sum of squared distances from the lines of sights. b)~$L_2$ triangulation method finds a 3D point which has the minimum sum of squared 2D distances between its projections and the corresponding points on images.}\label{fig:l2vsmid}
\end{figure*}
One of the fundamental tasks in 3D vision is reconstructing a point in 3D from its projections on the camera images in two or multiple views. This task, which is called triangulation, is used extensively in machine vision and robotics applications such as stereo vision \cite{tippetts2016review}, mapping \cite{castle2011wide} and \textit{structure-from-motion} \cite{bartoli2005structure,ramalingam2006generic}. It is obvious that a 3D point can be simply found by intersecting the lines of sight of each projection when no noise is present. But in practice, due to several sources of noise such as uncertainties in relative camera poses, errors in cameras intrinsic parameters, and subpixel inaccuracies in the position of matched points, all the lines do not necessarily intersect at one point or at all.
There are many attempts to solve the triangulation problem in the presence of uncertainties
\cite{hartley1997triangulation,lindstrom2010triangulation,lee2019closed}.

The common method for solving triangulation in two views, called the optimal method in the literature, is casting it as a nonlinear optimization problem. In such a problem, a new point is found as close as possible to the measured point in each view so that the lines of sights for the new points intersect. In \cite{hartley1997triangulation}, the authors compared several triangulation methods on several simulated datasets. The authors observed that the optimal triangulation method, which we call the $L_2$ method in this paper, outperforms other methods. Apart from good performance, the authors argued that the $L_2$ triangulation method has a nice property to be projective invariant. In the performed simulations, the authors considered the uncertainty in the position of corresponding points and not camera parameters.

In the majority of triangulation problems, there are uncertainties in both cameras' parameters and measured corresponding points.
Thus, a question arises here, ``Is the $L_2$ triangulation method still the best performing method?''
To the best of our knowledge, this question has not been investigated in the literature.
In this paper, we evaluate the performance of triangulation methods in the calibrated \textit{structure-from-motion} setting.
The calibrated \textit{structure-from-motion} after the work of \cite{nister2004efficient} has become the natural choice for \textit{structure-from-motion} applications.
In this case, we know cameras intrinsic parameters, but their extrinsic parameters as well as 3D points are estimated from the observed points in different views.
Knowing intrinsic calibration, one sees improvement over the accuracy and robustness of the structure and motion estimates \cite{nister2004efficient,kukelova2008polynomial}.

We will show that when uncertainty exists in camera extrinsic parameters, the $L_2$ triangulation method is no longer the state-of-the-art method. Interestingly, a simple mid-point method, i.e. the mid-point of lines of sights in different views, works much better in practice.  Fig.~\ref{fig:l2vsmid} depicts the difference between $L_2$ and mid-point triangulation methods.
The mid-point method not only gives better results but it can also be generalized to any number of views with no difficulties. The $L_2$ triangulation method is normally used for two views where the roots of a polynomial of degree $6$ needs to be computed \cite{hartley1997triangulation}. For more than two views, the method becomes computationally expensive and hard. For example, in the case of three views, the optimal solution is one of the real solutions among the set of $47$ general roots of a polynomial equation \cite{byrod2007improving}. The mid-point method in not projective or affine invariant \cite{hartley1997triangulation}, but this lack of invariance is not important for calibrated reconstruction.

Through extensive simulations, both on synthetic and real datasets, we have validated the high performance of the mid-point method.
The performance is defined as the accuracy of reconstruction where the optimal similarity transform is applied to the reconstructed 3D points.
%We have assessed the performance when we add uncertainty in relative camera poses.
We have assessed the performance when uncertainty in relative cameras poses exists.
We have also assessed the performance when uncertainty is caused by commonly used \textit{structure-from-motion} approach, where first the essential matrix\footnote{The essential matrix corresponding to a pair of cameras with relative orientation, $R$, and translation, $t$, is defined as $E=[t]_\times R$.} is estimated from point correspondences and then relative poses are estimated \cite{nister2004efficient} and finally the 3D structure is obtained. In all of these experiments, we see that the mid-point method outperforms other approaches. Thus, we suggest that unlike the common practice, the mid-point method should be used in \textit{structure-from-motion} applications.

\section{Related Works}
The $L_2$ triangulation approach, which is known as the optimal method, finds the 3D point that minimizes the $L_2$ reprojection errors in the image domain \cite{kanatani1996statistical}.
This leads to finding the optimal, maximum-likelihood, solution under the assumption of Gaussian noise in the position of projections.
\cite{hartley1997triangulation} showed that minimizing the $L_2$ reprojection errors, for the case of two images, can be reduced to finding the stationary points of $6^{th}$ degree polynomial and selecting the best by evaluating the objective function.
A Gr\"{o}bner basis based algorithm for minimizing the $L_2$ reprojection errors, in the case of three image observations, is proposed by \cite{stewenius2005hard}. They showed that the optimal solution is one of the real solutions among a set of $47$ general roots of a certain polynomial equation. Since their approach has a significant computational cost, an alternative method of \cite{byrod2007improving} can be used, which presents techniques that improves the numerical stability of Gr\"{o}bner basis solvers and significantly reduces the computational costs.

Because of the non-convexity and complexity of solving the $L_2$ norm \cite{hartley2013verifying}, other cost functions were considered in the literature.
For instance, a choice which is robust to outliers is to minimize the $L_1$ reprojection errors. \cite{hartley1997triangulation} find the $L_1$ optimal solution in closed form by solving a polynomial of degree~$8$.
They also state that the $L_1$ optimization has slightly less 3D error than the $L_2$ optimization in real experiments.
%\cite{hartley2004sub} showed that the triangulation with $L_1$ norm, for any number of views, is a convex optimization problem.
Another popular approach is to find $L_\infty$ answer which is optimal under the assumption of uniform noise \cite{hartley2007optimal,hartley2004sub,li2007practical,olsson2010outlier,sim2006removing}.
%6 9 16 26 30 of Closed-form...
Angular errors were studied in \cite{lee2019closed} and closed-form optimal solutions were derived for $L_1$, $L_2$, and $L_\infty$ angular errors.

In this paper the accuracy of different triangulation approaches in a calibrated \textit{structure-from-motion} process \cite{toldo2015hierarchical,ramalingam2006generic} is investigated. It is shown that the 3D baseline triangulation approach has less sensitivity to uncertainties in cameras extrinsic parameters and also has more accuracy in 3D reconstruction in a full \textit{structure-from-motion} process where positions and orientations of the cameras are estimated from observations.

\section{Preliminaries}
\subsection{Camera Model and Parameters}
Let $\vu$ be the projection of a point~${\vp}$ on a camera's image plane. The projection is obtained by $ \vu=P[{\vp};1]$, where $P$ is the camera matrix. The camera matrix $P$ is given by
$P= K[R~|-R\vc]$
, where $K$ is the camera calibration matrix, and $R$ and $\vc$ are the orientation and position of the camera with respect to a world coordinate system. The line of sight of the camera image is the line that passes through the camera point $\vc$ towards direction $R^{-1}K^{-1}{\vu}$.
\subsection{Optimal Triangulation}
2D baseline triangulation, which is known as the optimal triangulation in the literature, finds a $3D$ point $\vp$ so that its projected points on the cameras, $\hat{\vu}_i, i\in\{1,\ldots,N_c\}$, have the minimum sum Euclidean distance from measurements $\vu_i$s. Hence, it minimizes the following cost function:
\begin{equation}\label{optimal_cost}
    f(\vp) = \sum_{i=1}^{N_c}d(\hat{\vu}_i,\vu_i)^2,\quad  \hat{\vu}_i=P[{\vp};1],\ i\in\{1,\ldots,N_c\},
\end{equation}
in which $d(\hat{\vu}_i,\vu_i)$ is the Euclidean distance between the projected point and its measurement in the $i^{th}$ image.
Assuming independent Gaussian noise in the image domain and known cameras positions and orientations, this method provides the maximum-likelihood estimation of 3D points.
%As it shown in the literature, this method provides the maximum-likelihood estimation, where the camera positions and orientations are known and the noise is independent Gaussian in the image domain.
%The independent Gaussian noise is a reasonable assumption, but in real experiments there are also uncertainties in the position and orientations of cameras. Thus there arises a question that is ``optimal triangulation'' still the best choice?

\subsection{Mid-Point Triangulation}
Another simple triangulation method is to find a $3D$ point $\vp$ that minimizes $3D$ distances from the lines of sights.
The goal of this method is to minimize the following cost function:
\begin{equation}\label{tri3d_cost}
    f(\vp) = \sum_{i=1}^{N_c}d(\vp,\vl_i)^2,
\end{equation}
in which $\vl_i$s are the lines of sights and $d(\vp,\vl_i)$ is the distance between $\vp$ and $\vl_i$. For any number of cameras, minimizing \eqref{tri3d_cost} is a linear least squares problem and can be solved in a closed form.
%this method has the following closed form solution.
%\begin{equation}\label{optimal_cost}
%    X = \frac{\sum_{i=1}^{N_c}d(X,\vl_i)^2}{}
%\end{equation}

\subsection{Accuracy of the reconstruction}
\begin{figure}
    \centering
    \includegraphics[width=0.48\textwidth,clip,keepaspectratio]{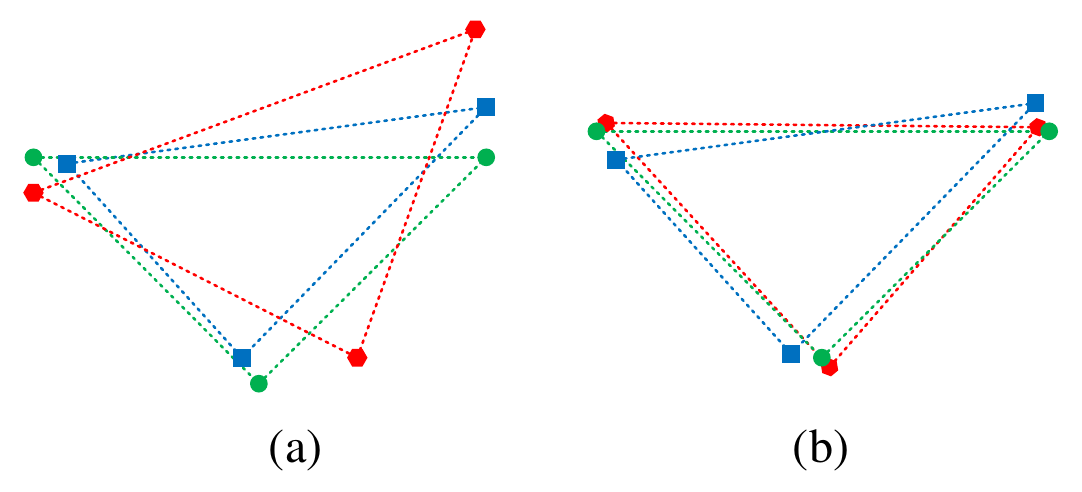}
    \caption{The ground truth point cloud (green circles) and the estimated point clouds (blue squares and red hexagons). As it is obvious in (a), the blue points have a smaller sum squared distance to the ground truth versus the red points. But after finding the best rotation, translation, and scale for both red and blue points, done for (b), it reveals that the red points are the better estimation of the ground truth green points.}\label{fig:groundtruth}
\end{figure}

Point cloud reconstructed by a \textit{structure-from-motion} procedure is obtained up to a scaled Euclidean transformation %, i.e. a translation and a rotation.
(a more general projective ambiguity exists in the uncalibrated approach).
Suppose that $\hat{\vp}_i, i\in\{1,\ldots,N\}$ are estimated points, %resulting from different procedures,
and $\vp_i$s are the ground truth (with known correspondences).
%Suppose that we have an estimated point cloud $\hat{\vp}_i, i\in\{1,\ldots,N\}$, %resulting from different procedures,
%and the ground truth $\vp_i$s (with known correspondences).
As shown in Fig.~\ref{fig:groundtruth}, the accuracy of the estimation is obtained by finding a scaled Euclidean transformation such that the estimated point cloud are aligned to the ground truth as much as possible. Mathematically speaking
\begin{equation}\label{rts}
    \min_{R,\vt,s} \sum_{i=1}^{N}d\bigl(sR\hat{\vp}_i+\vt,\vp_i\bigr)^2,
\end{equation}
where $R$, $\vt$, and $s$ are the rotation matrix, translation, and scale parameters of the scaled Euclidean transformation.

\subsection{Reconstruction in a \textit{Structure-from-Motion} Process}
A typical \textit{structure-from-motion} framework \cite{ramalingam2006generic,schonberger2016structure,sweeney2015optimizing} comprises the following steps:
\begin{enumerate}
    \item Pairwise images registration
    \begin{itemize}
        \item Feature extraction and matching \cite{lowe1999object,bay2008speeded,lo2009local}
        \item Finding relative rotations and translations between all pairs of images with the matched features \cite{torr2000mlesac,nister2004efficient,kukelova2008polynomial}
    \end{itemize}
    \item Camera pose estimation
    \begin{itemize}
        \item Solving the viewing graph created by the pairwise image registrations to find camera positions and orientations \cite{ozyesil2015robust,sweeney2015optimizing,jiang2013global,zhu2018very,hartley2013rotation,chatterjee2017robust,arrigoni2018robust}
    \end{itemize}
    \item Triangulation
    \begin{itemize}
        \item Reconstructing 3D points by triangulating corresponding points.
    \end{itemize}
    The obtained camera poses and 3D reconstructed points are usually refined by a \textit{bundle adjustment} step.
    %Regarding the subject of the paper, i.e. triangulation, we refer to the three mentioned steps as the reconstruction process and exclude the bundle adjustment step.
    {\color{black}In this paper, we only consider up to the triangulation adjustment and exclude the bundle adjustment step.}
    %In this paper, for the sake of simplicity, we do not include the bundle adjustment step in the reconstruction process.

    %This paper compares different triangulation methods
    %Since the purpose of this paper is comparing different triangulation methods, we
    %As we aim to
\end{enumerate}

\section{Experiments}
The experiments are two-fold. First, the sensitivity of different triangulation methods to the error in camera poses are compared. Then, the accuracy of different triangulation methods are evaluated in a full reconstruction procedure on synthetic and real datasets.

\subsection{Sensitivities}
%As it shown in the literature, the optimal triangulation is the maximum likelihood estimation where the camera positions and orientations are known and the noise is independent Gaussian in the image domain. The independent Gaussian noise is a reasonable assumption, but in real experiments there are also uncertainties in the position and orientations of cameras. Thus there arises a question that is ``optimal triangulation'' still the best choice?

In this part, the mid-point method is compared to the $L_2$ method \cite{hartley1997triangulation}% and some other common triangulation methods, namely
, iteratively reweighted mid-point \cite{yang2019iteratively}, minimizing $L_1$ reprojection errors \cite{hartley1997triangulation}, and minimizing $L_1$ and $L_2$ angular errors \cite{lee2019closed}.%in the following aspects:
The comparison criteria are:
%The sensitivity of the following errors to the position and orientation of the cameras are obtained and compared for different triangulation methods.
%The following sensitivities to the error in position and orientation of cameras are obtained and compared for different triangulation methods.
%\begin{itemize}
%  \item First the sensitivity of error of a single point triangulation.
%  \item Second the sensitivity of error in the distance of two triangulated points.
%  \item Third the sensitivity of error in the angle of three triangulated points.
%\end{itemize}
\begin{itemize}
  \item Position error sensitivity: The sensitivity of the error in the position of a single triangulated point.
  \item Distance error sensitivity: The sensitivity of the error in the distance between two triangulated points.
  \item Angle error sensitivity: The sensitivity of the error in the angles of a triangle composed of three triangulated points.
\end{itemize}

To evaluate the mentioned sensitivities for different triangulation methods, three configuration %for the position and orientation of camerat
are considered for two-views triangulation:
\begin{enumerate}
  \item [] Conf.~1)~ $\vc_1=[-5,-1,0]^T, \vc_2=[-5,+1,0]^T$ and the both cameras point at origin.
  \item [] Conf.~2)~ $\vc_1=[-12,0,0]^T, \vc_2=[-2,0,0]^T$ and the both cameras point at origin.
  \item [] Conf.~3)~ $\vc_1=[-10,2,-1]^T, \vc_2=[-5,-2,1]^T$ and the both cameras baselines are aligned with the global coordinate $x$ direction.
\end{enumerate}

The first two configurations are the same as the two configurations in \cite{hartley1997triangulation}.
In fact, the first configuration simulates a camera moving forward and looking straight ahead and the second configuration simulates an aerial imaging procedure. Since these two configurations are special cases, a more general configuration is added to them.

\subsubsection{Position Error Sensitivity}
In this part the sensitivity of $3D$ error of different triangulation methods to errors in positions and orientations of cameras are evaluated. For each configuration, a point $\vp$ is placed in an sphere centered at the origin with the diameter of $0.5$, and the projected points on the cameras are obtained. The positions of the cameras are perturbed by random Gaussian noise vectors and the orientation of the cameras are perturbed by random rotations with random axes and Gaussian random angles. The reconstructed point $\hat \vp$, is obtained by different triangulation methods. Euclidean distance between points is used for computing $3D$ error:
\begin{equation}\label{err_t}
    e = d(\hat\vp,\vp).
\end{equation}

\begin{figure}
    \centering
    \includegraphics[width=0.48\textwidth,clip,keepaspectratio]{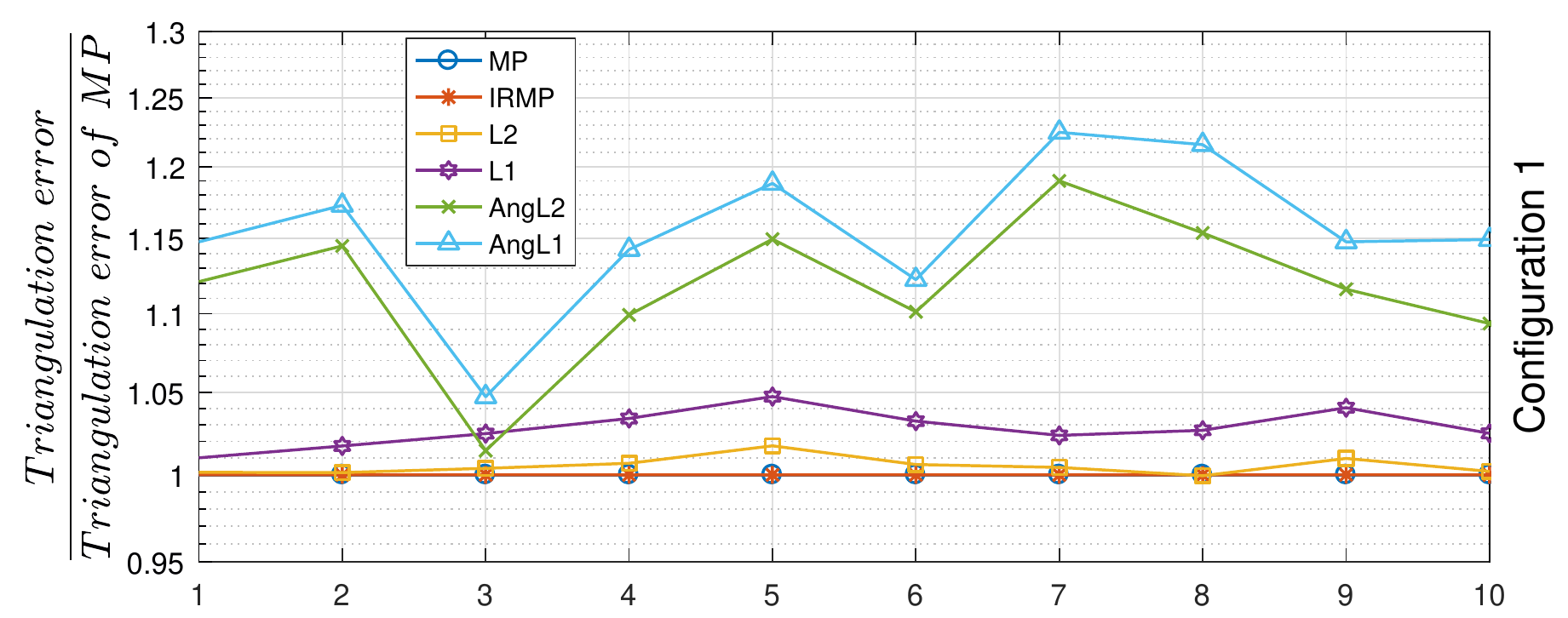}
%    \caption{Position sensitivity Conf.1}\label{fig:Exp1C1_mt}
%\end{figure}
%\begin{figure}
%    \centering
    \includegraphics[width=0.48\textwidth,clip,keepaspectratio]{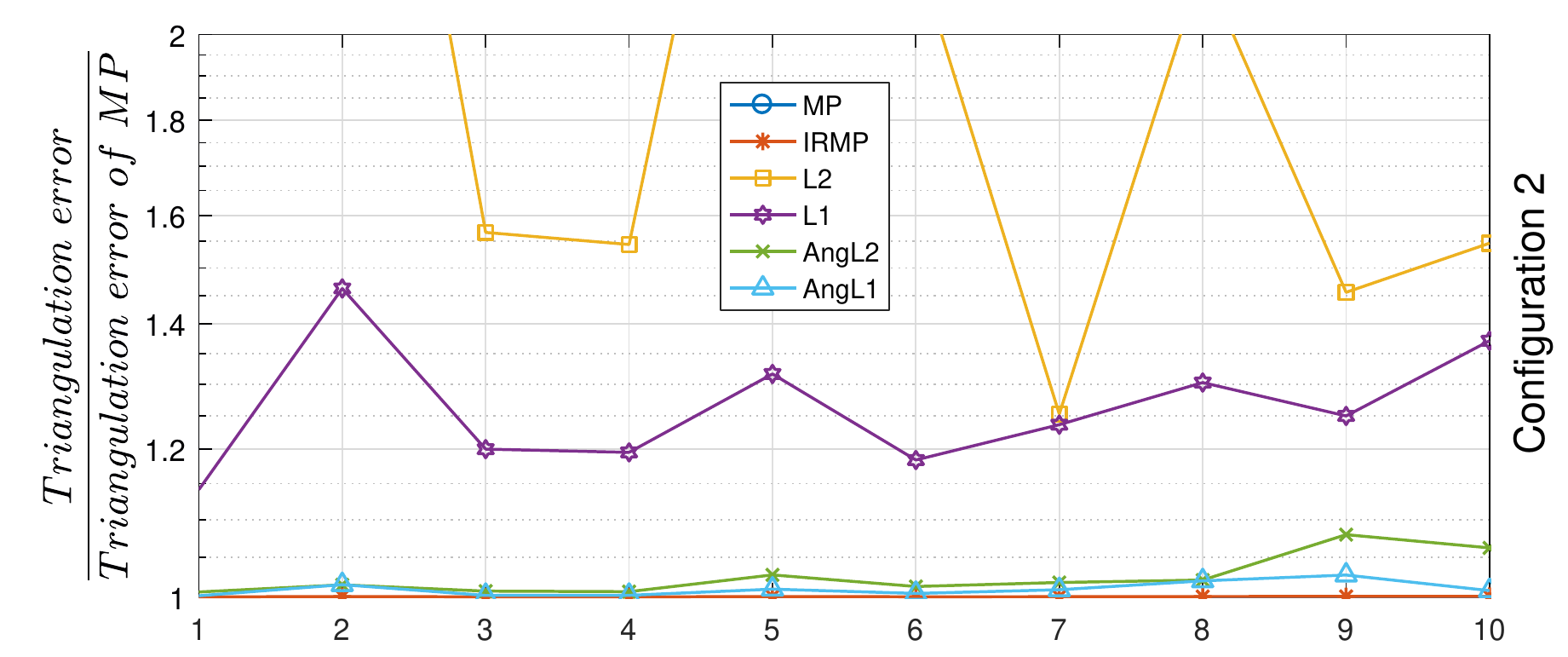}
%    \caption{Position sensitivity Conf.2}\label{fig:Exp1C2_mt}
%\end{figure}
%\begin{figure}
%    \centering
    \includegraphics[width=0.48\textwidth,clip,keepaspectratio]{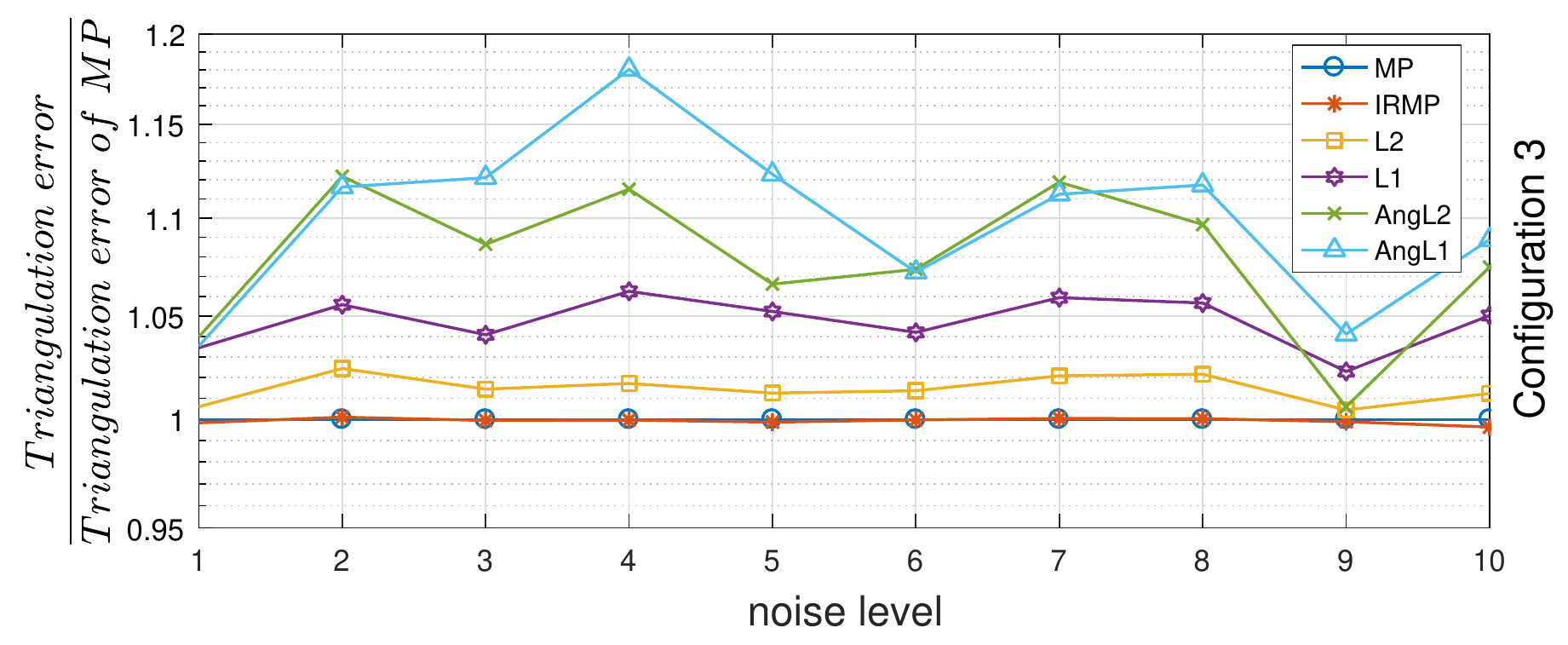}
%    \caption{Position sensitivity Conf.3}\label{fig:Exp1C3_mt}
    \caption{The position error \eqref{err_t} sensitivity of different methods in - from top to bottom - configurations 1, 2, 3. }\label{fig:Exp1_mt}
\end{figure}

This procedure repeated $100$ times for each configuration and for each noise level.
The standard deviation of the position Gaussian noise is $0.01$, and the standard deviation of the angle Gaussian noise is $0.1$ degree for the noise level $1$. The standard deviations are multiplied by the noise levels.
%Figures~\ref{fig:Exp1C1_mt}, \ref{fig:Exp1C2_mt}, and \ref{fig:Exp1C3_mt}
Fig.~\ref{fig:Exp1_mt} shows the mean error of different triangulation methods in confs.~1,~2, and~3. %, respectively.

\subsubsection{Distance Error Sensitivity}
To evaluate the distance error sensitivity of different methods, two points $\vp_1$ and $\vp_2$ are randomly placed in the sphere of the previous part. The projections are computed, cameras positions and orientations are perturbed by the noise, and different triangulation methods are applied to find two estimated $3D$ points $\hat \vp_1$ and $\hat \vp_2$. The error is the absolute value of difference of the distance between $\vp_1$ and $\vp_2$, and the distance between $\hat \vp_1$ and $\hat \vp_2$:
\begin{equation}\label{err_d}
    e = \left|d(\vp_1,\vp_2)-d(\hat \vp_1,\hat \vp_2)\right|.
\end{equation}

\begin{figure}
    \centering
    \includegraphics[width=0.48\textwidth,clip,keepaspectratio]{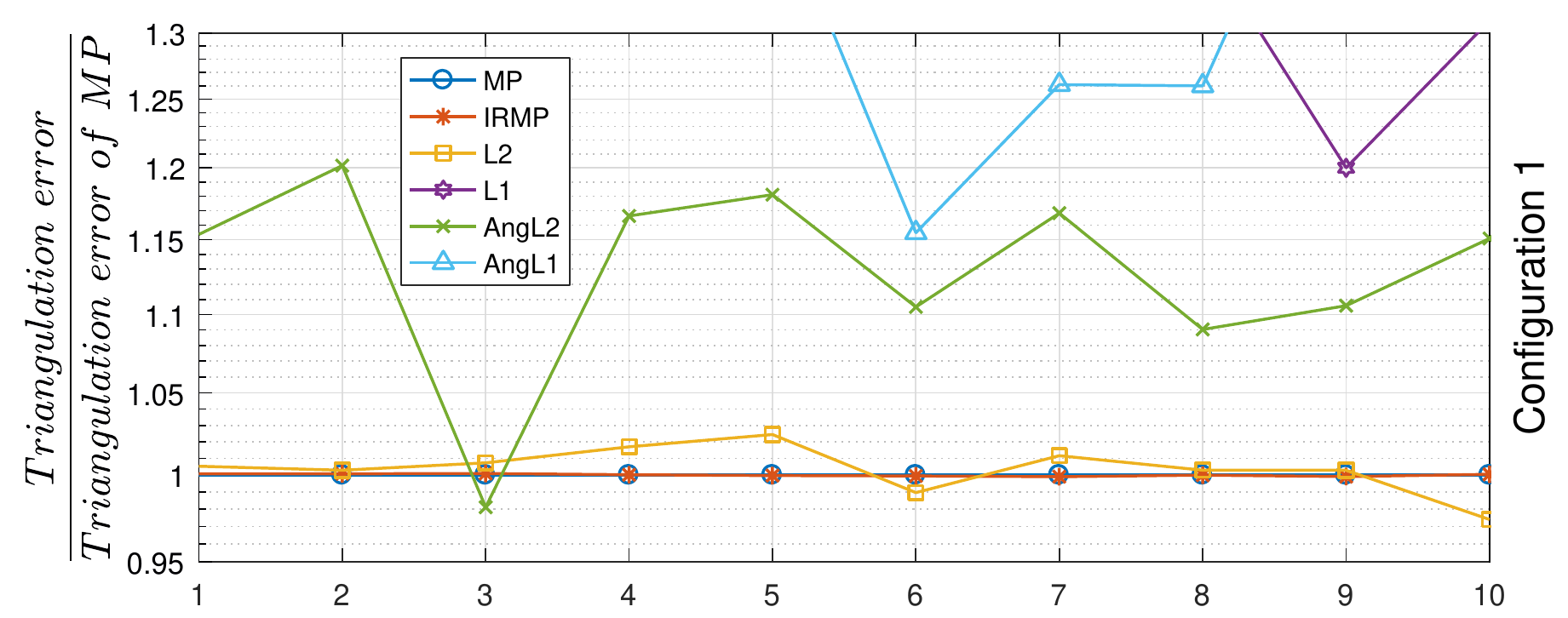}
%    \caption{Distance sensitivity Conf.1}\label{fig:Exp1C1_md}
%\end{figure}
%\begin{figure}
%    \centering
    \includegraphics[width=0.48\textwidth,clip,keepaspectratio]{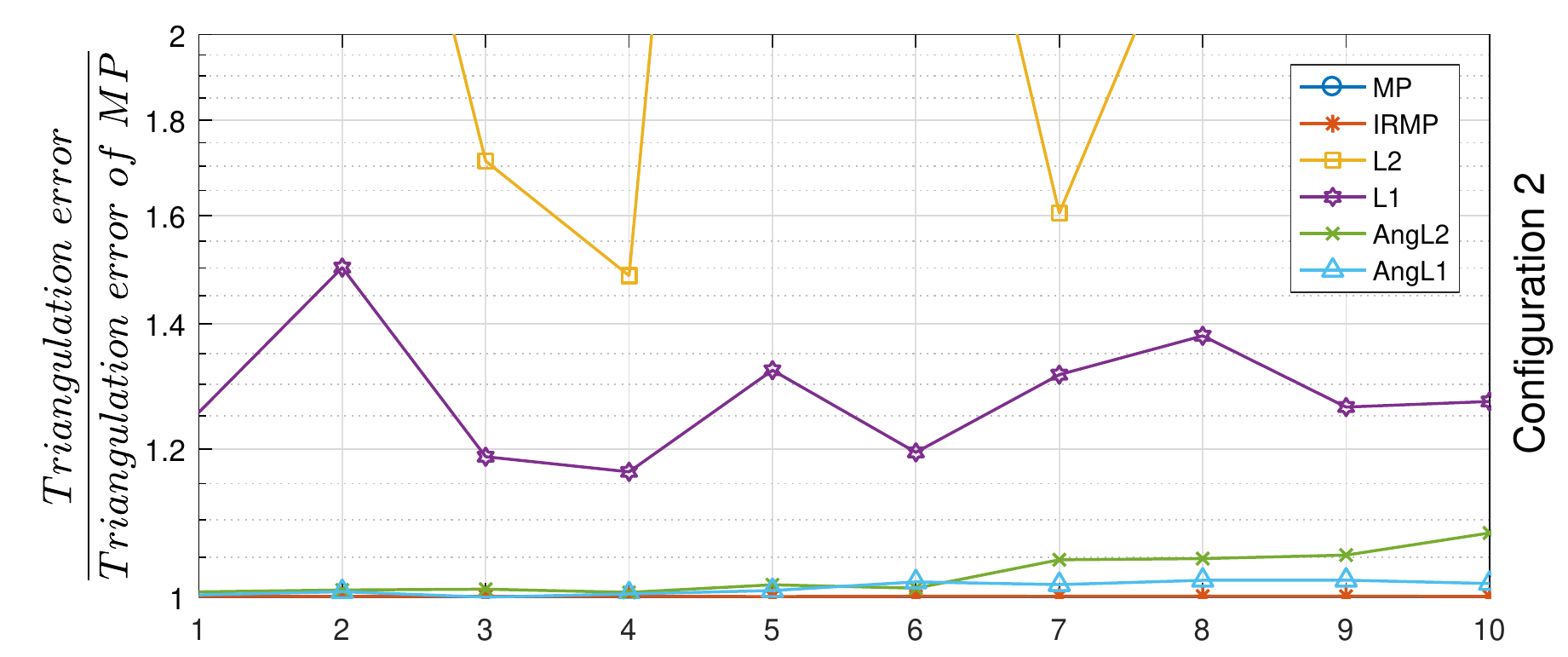}
%    \caption{Distance sensitivity Conf.2}\label{fig:Exp1C2_md}
%\end{figure}
%\begin{figure}
%    \centering
    \includegraphics[width=0.48\textwidth,clip,keepaspectratio]{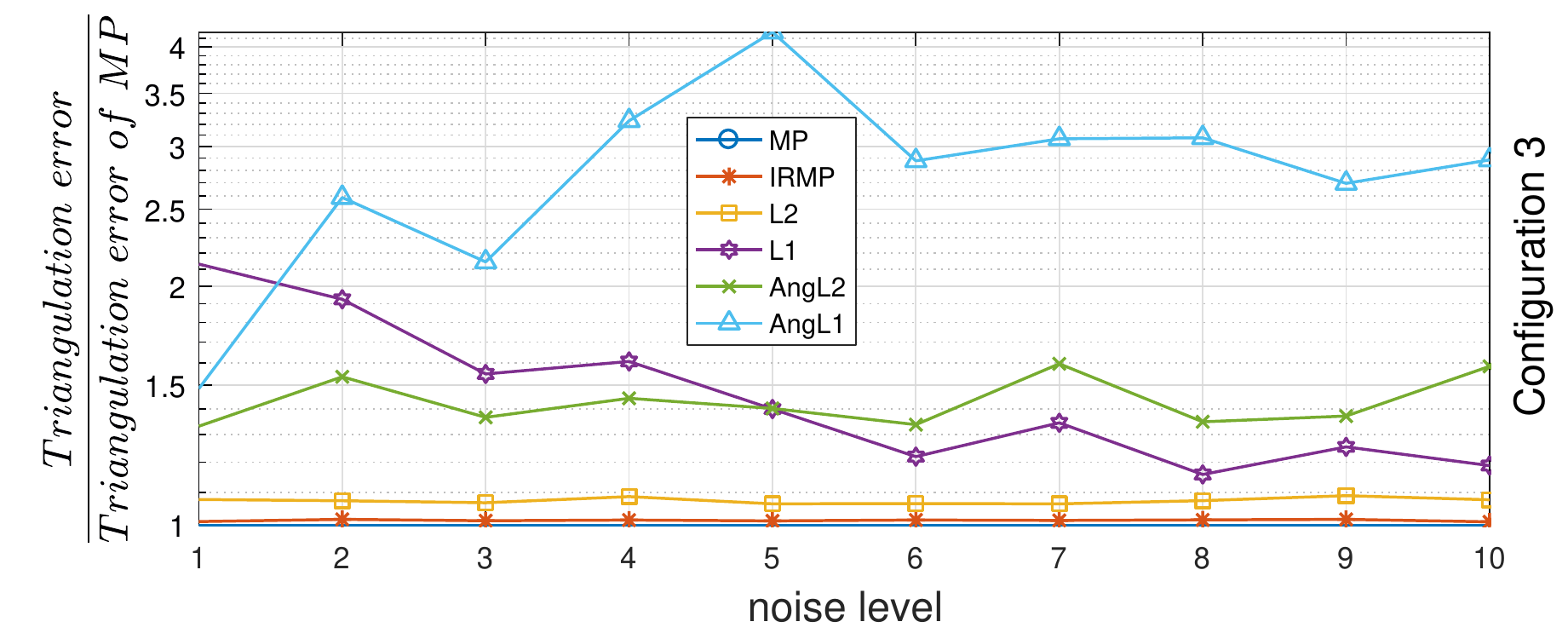}
%    \caption{Distance sensitivity Conf.3}\label{fig:Exp1C3_md}
    \caption{The distance error \eqref{err_d} sensitivity of different methods in - from top to bottom - Configurations 1, 2, 3. }\label{fig:Exp1_md}
\end{figure}

This procedure is repeated $100$ times for each configuration and for each noise level. The mean error of different triangulation methods are shown in Fig.~\ref{fig:Exp1_md}.
%figures~\ref{fig:Exp1C1_md}, \ref{fig:Exp1C2_md}, and \ref{fig:Exp1C3_md}.

\subsubsection{Angle Error Sensitivity}
To compute the angle error sensitivity of different methods, three points $\vp_1$, $\vp_2$, and $\vp_3$ are randomly placed in the aforementioned sphere and the projections are computed. The positions and orientations of the cameras are perturbed by the noise and three $3D$ points $\hat \vp_1$, $\hat \vp_2$, $\hat \vp_3$ are estimated by different triangulation methods. The error is defined as the absolute value of difference between the angle between two vectors $\vp_2-\vp_1$ and $\vp_3-\vp_1$, and the angle between two vectors $\hat \vp_2-\hat \vp_1$ and $\hat \vp_3-\hat \vp_1$:
\begin{equation}\label{err_a}
    e = \left|\measuredangle(\hat \vp_2-\hat \vp_1,\hat \vp_3-\hat \vp_1)-\measuredangle(\vp_2-\vp_1,\vp_3-\vp_1)\right|.
\end{equation}

\begin{figure}
    \centering
    \includegraphics[width=0.48\textwidth,clip,keepaspectratio]{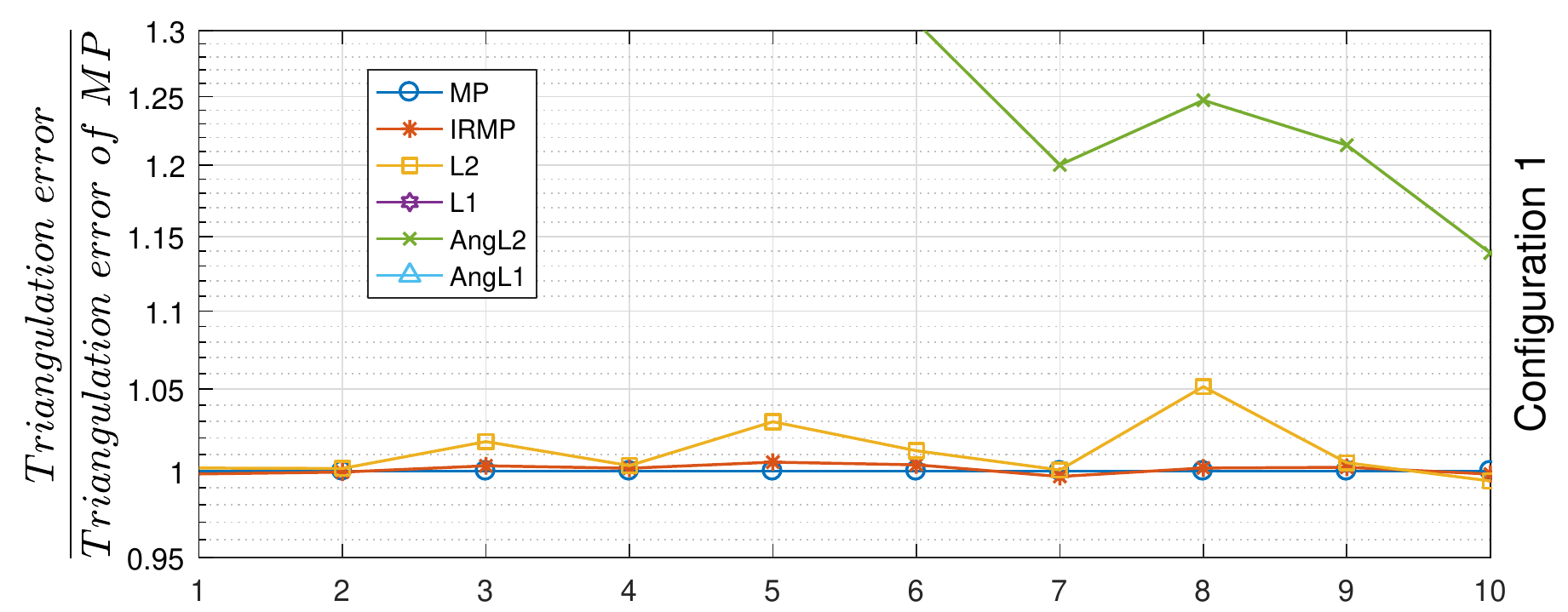}
%    \caption{Angle sensitivity Conf.1}\label{fig:Exp1C1_ma}
%\end{figure}
%\begin{figure}
%    \centering
    \includegraphics[width=0.48\textwidth,clip,keepaspectratio]{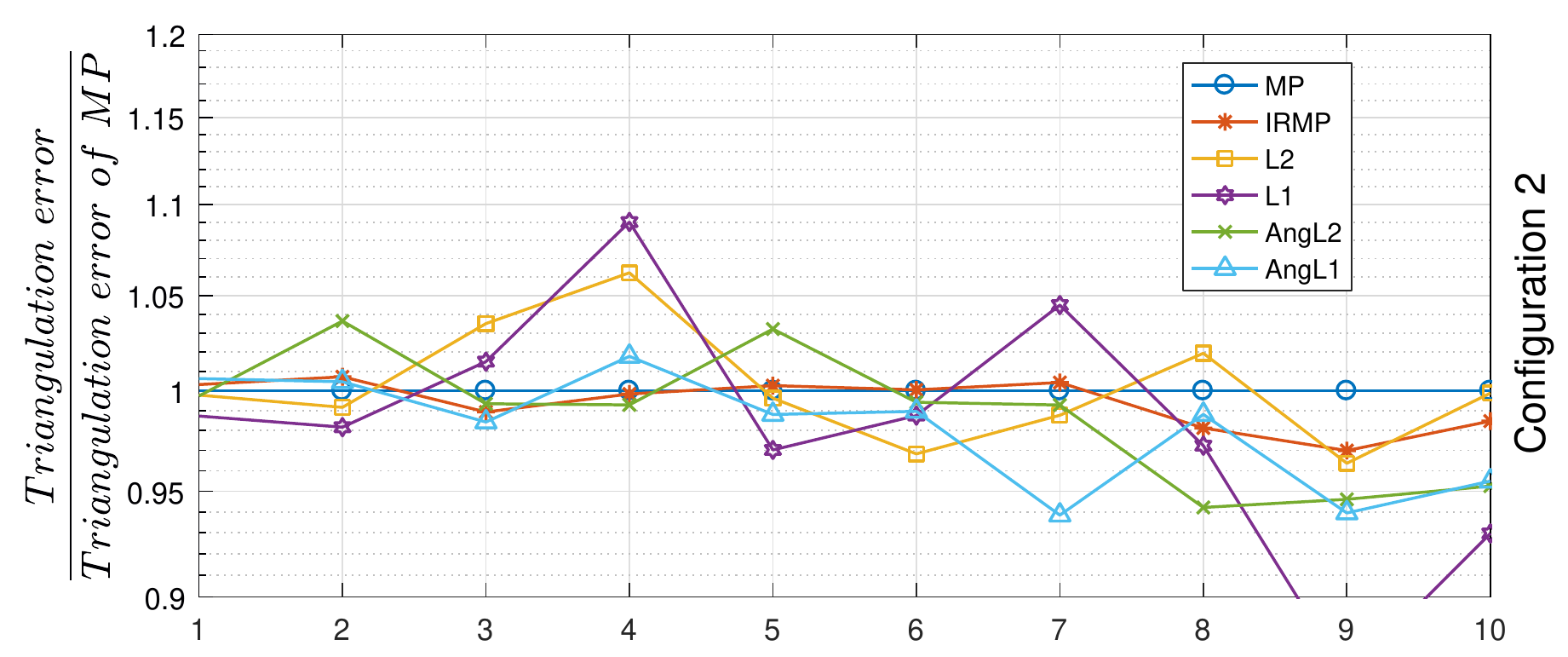}
%    \caption{Angle sensitivity Conf.2}\label{fig:Exp1C2_ma}
%\end{figure}
%\begin{figure}
%    \centering
    \includegraphics[width=0.48\textwidth,clip,keepaspectratio]{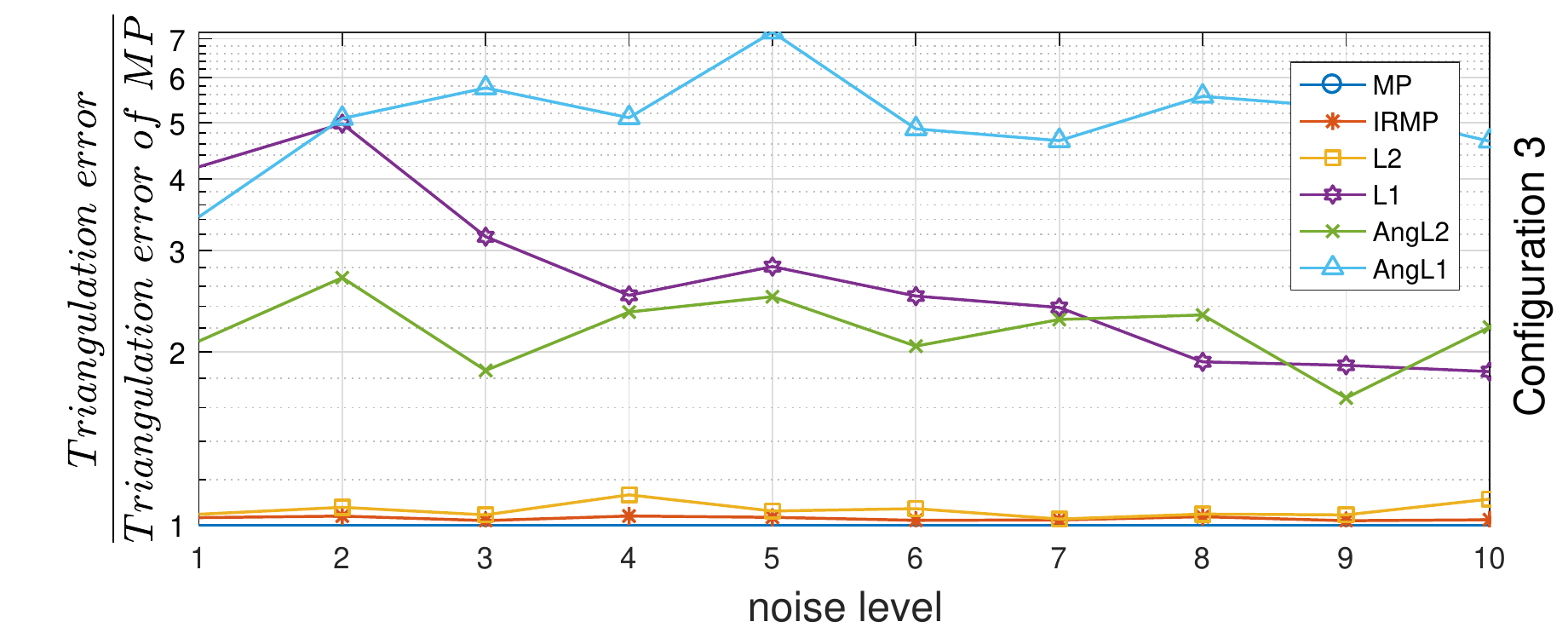}
%    \caption{Angle sensitivity Conf.3}\label{fig:Exp1C3_ma}
    \caption{The angle error \eqref{err_a} sensitivity of different methods in - from top to bottom - Configurations 1, 2, 3. }\label{fig:Exp1_ma}
\end{figure}

This procedure is repeated $100$ times for each configuration and for each noise level. The mean error of different triangulation methods are shown in Fig.~\ref{fig:Exp1_ma}.
%figures~\ref{fig:Exp1C1_ma}, \ref{fig:Exp1C2_ma}, and \ref{fig:Exp1C3_ma}.

From these experiments, it can be concluded that the mid-point method and its variant are the best performing methods when there is uncertainty in the cameras parameters. In reality, there are uncertainty in both cameras parameters and image points. This is addressed in the following subsection.

\subsection{Full Reconstruction Procedure on Synthetic Datasets}
%In the following experiments, we assess the performance of different methods in a full structure-from-motion reconstruction procedure where there are uncertainties in image points and consequently this leads to uncertainty in cameras parameters. We first evaluate the performance in the case of two cameras, where we compute camera matrices with the essential matrix. Then, we evaluate the performance for the case of more than two cameras, where an additional pose graph optimization is needed to compute the camera matrices.
In the following experiments, the performance of the above methods in a full structure-from-motion reconstruction procedure is assessed. The datasets have uncertainties in image points and consequently there is uncertainty in cameras extrinsic parameters. In this section, first the performance in the case of two cameras is evaluated, where camera poses are computed through the essential matrix. Then, the performance for the case of more than two cameras are evaluated, where an additional viewing graph optimization is needed to be solved for computing the camera poses.
\subsubsection{Two Cameras}
\begin{figure}
    \centering
    \includegraphics[width=0.48\textwidth,clip,keepaspectratio]{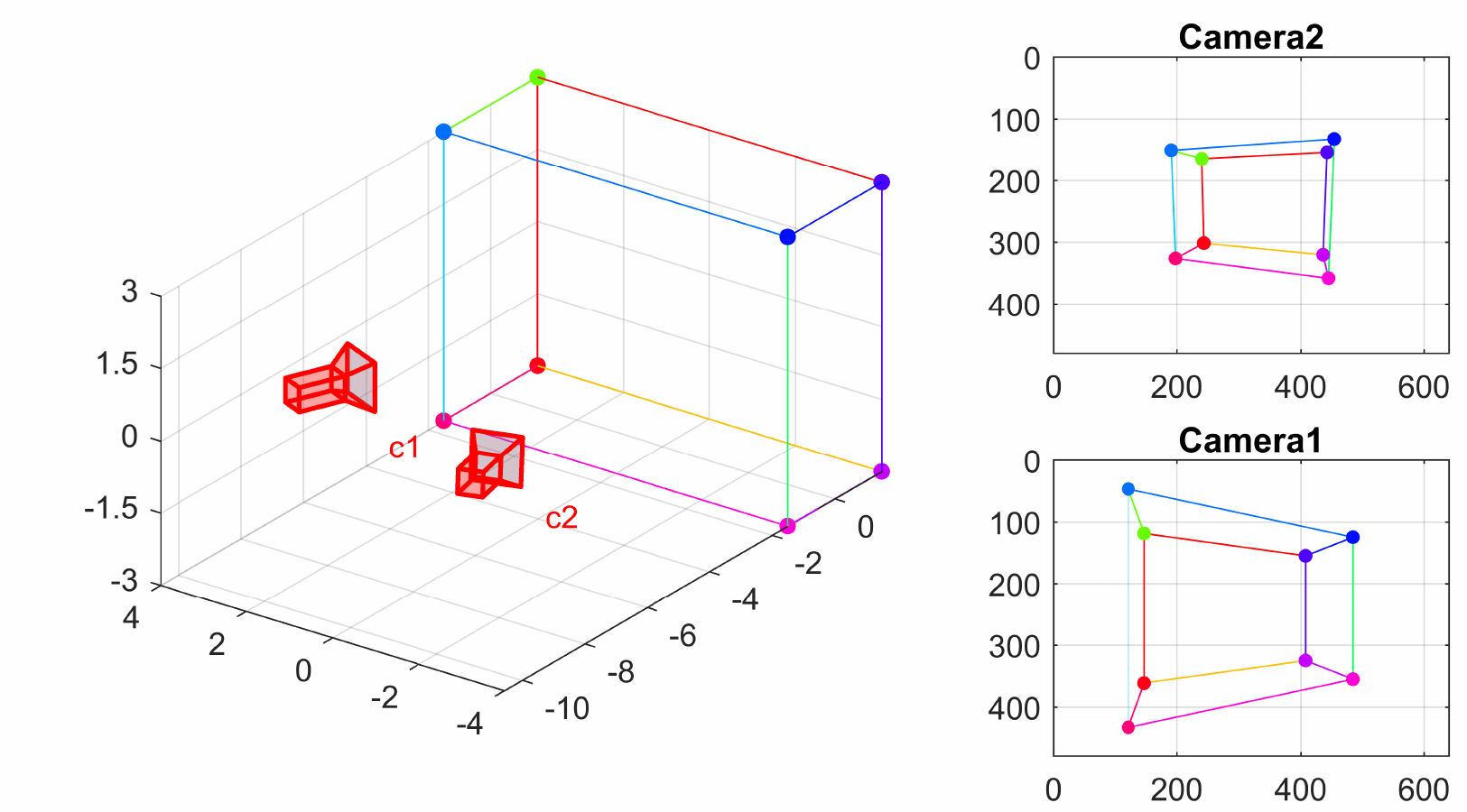}
    \caption{The configuration of the cameras and the box, and the images of the box on the cameras. The cameras are placed at $\vc_1=[-7,3,0]^T, \vc_2=[-10,-3,1]^T$. Both cameras looking at the origin. The dimensions of the box are $3\times8\times6$ and, it is centered at the origin.}\label{fig:Conf4}
\end{figure}

To evaluate the performance of different triangulation methods for the case of two cameras, the following steps are done on the simulation setup of Fig.~\ref{fig:Conf4}:
\begin{enumerate}
  \item $N=20$ points are randomly selected in a box.
  \item The points are projected on the cameras.
  \item The projections are displaced by an unbiased Gaussian random noises with the standard deviation of one pixel.
  \item The essential matrix between the two cameras is estimated by the method of \cite{kukelova2008polynomial}.
  \item The relative rotation and translation are estimated.
  \item The cameras poses are calculated from the relative observations.
  \item The corresponding $3D$ points are reconstructed by triangulation.
  \item The best rotation, translation, and scale that makes the $3D$ triangulated points match the selected points in the box are obtained and the errors are computed.
\end{enumerate}

%The cameras and the box are configured as shown in Fig.~\ref{fig:Conf4}.
The procedure is repeated $100$ times.
The cameras have the same calibration matrix
\begin{equation}\label{K}
    K =
    \begin{bmatrix}
      300 & 0   & 320 \\
      0   & 300 & 240 \\
      0   & 0   & 1
    \end{bmatrix},
\end{equation}
and the images have $640\times480$ pixels.

\begin{figure}
    \centering
    \includegraphics[width=0.5\textwidth,clip,keepaspectratio]{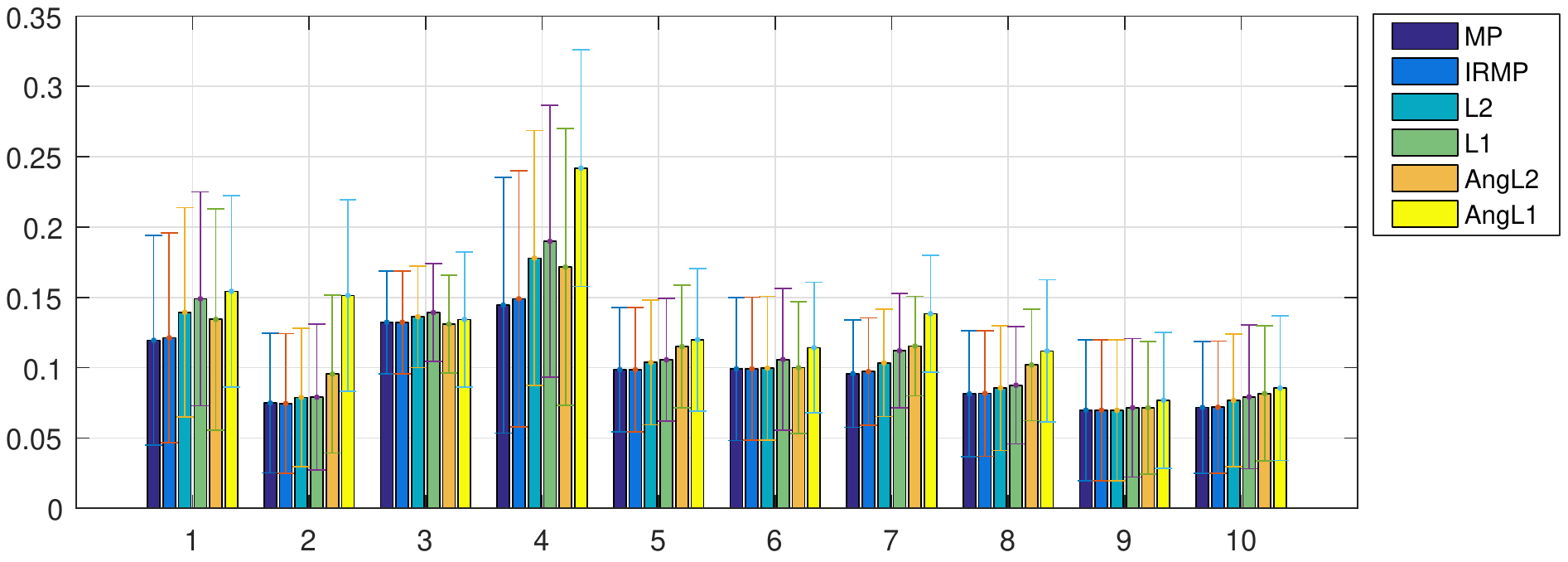}
    \caption{The mean and standard deviation of 3D triangulation errors for different methods in the first ten random runs on the synthetic dataset. The errors are computed for random 20 points selected in a region. The cameras and the region containing the points are shown in Fig.~\ref{fig:Conf4}. %- $N_c = 2$ - 10 first run - mean and std 3D error of 20 points
    }\label{fig:ExpF10}
\end{figure}

\begin{figure}[t]
    \centering
    \includegraphics[width=0.45\textwidth,clip,keepaspectratio]{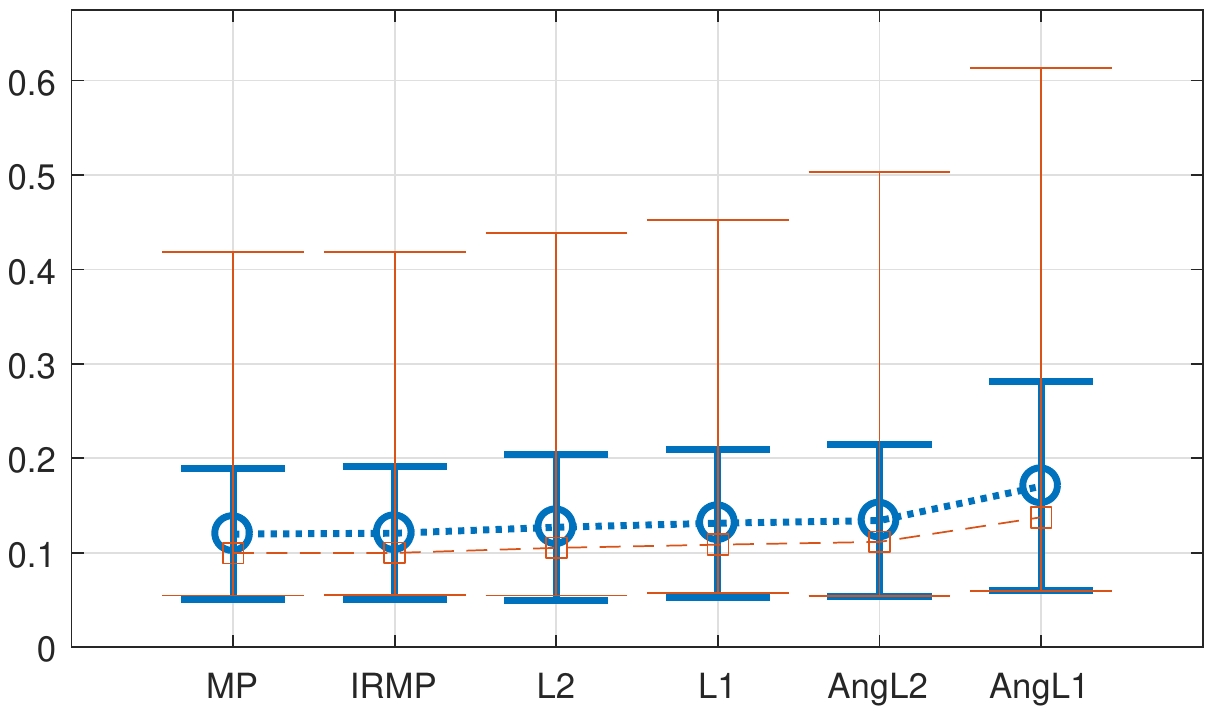}
    \caption{The mean, standard deviation, median, minimum, and maximum of 100 mean 3D triangulation error for different methods. The errors are computed for random 20 points selected in a region. The cameras and the region containing the points are shown in Fig.~\ref{fig:Conf4}.% - $N_c = 2$ - 100 run - mean, std, median, min and max of mean 3D error
    }\label{fig:ExpF100}
\end{figure}

Fig.~\ref{fig:ExpF10} shows the mean and standard deviation of triangulation errors of all $20$ points for different methods in the first~$10$ experiments. Fig.~\ref{fig:ExpF100} shows the mean, median, standard deviation, minimum, and maximum of the mean error of all $100$ experiments.

\subsubsection{More Than Two Cameras}\label{More Than Two Cameras}
If $N_c$ cameras ($N_c>2$) are involved in the reconstruction process, the essential matrices and consequently the relative positions and orientations are computed for every $2$-combinations of $N_c$ cameras. The $\binom{N_c}{2}$ relative observations of orientations and directions create a viewing graph which should be solved to estimate the cameras poses. In the $4^{th}$ step of the reconstruction procedure, the essential matrices are estimated for any $2$-combinations of $N_c$ cameras, and in $5^{th}$ step, a viewing graph is created from relative measurements. Obtained viewing graph is solved in step $6$ to estimate the positions and orientations of the cameras.

\begin{figure}[t]
    \centering
    \includegraphics[width=0.48\textwidth,clip,keepaspectratio]{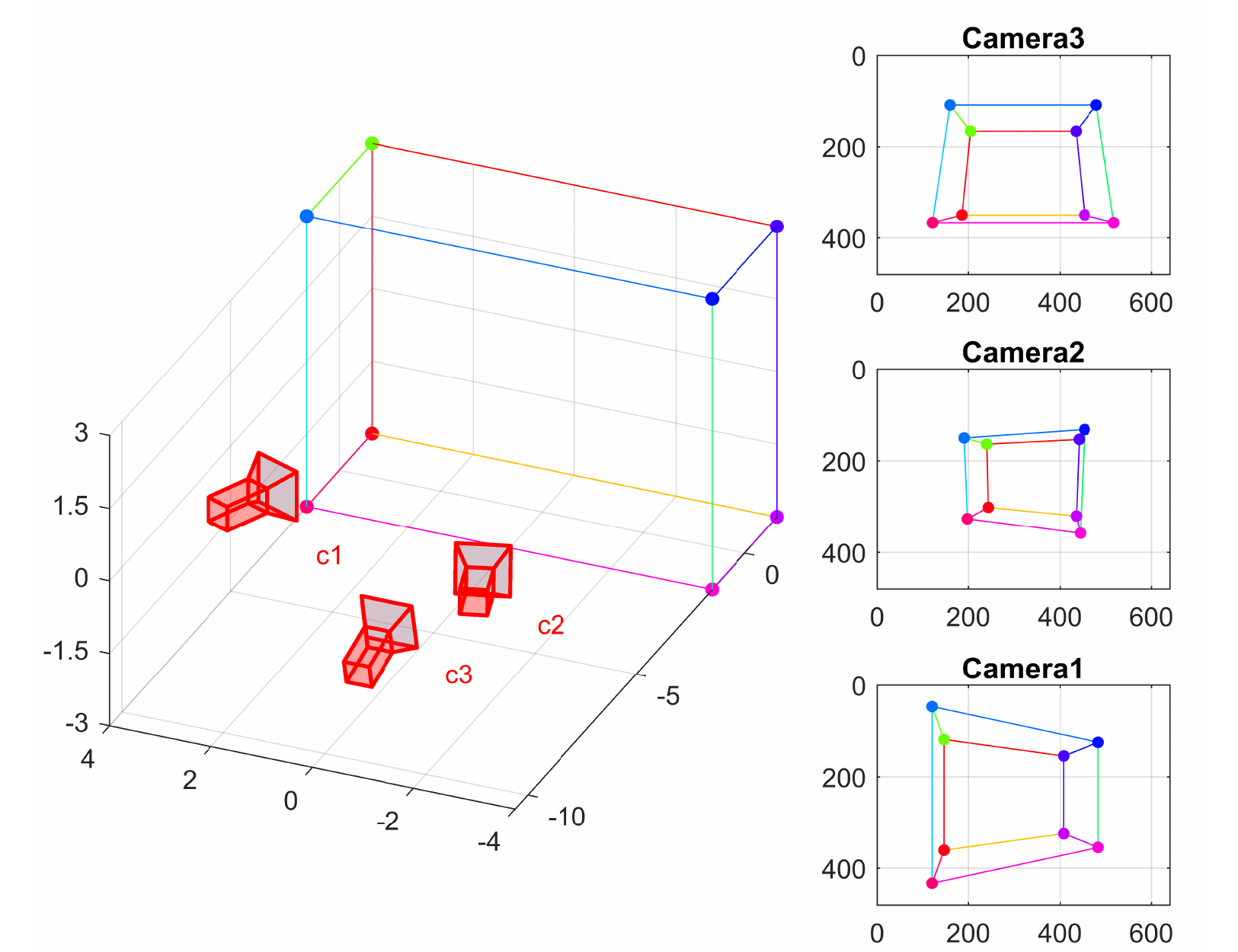}
    \caption{The configuration of the box and cameras $\vc_1$ and $\vc_2$ are the same as Fig.~\ref{fig:Conf4}, and the third camera is placed at $\vc_3=[-8,0,-2]^T$.}\label{fig:Conf4NC3}
\end{figure}

\begin{figure}
    \centering
    \includegraphics[width=0.5\textwidth,clip,keepaspectratio]{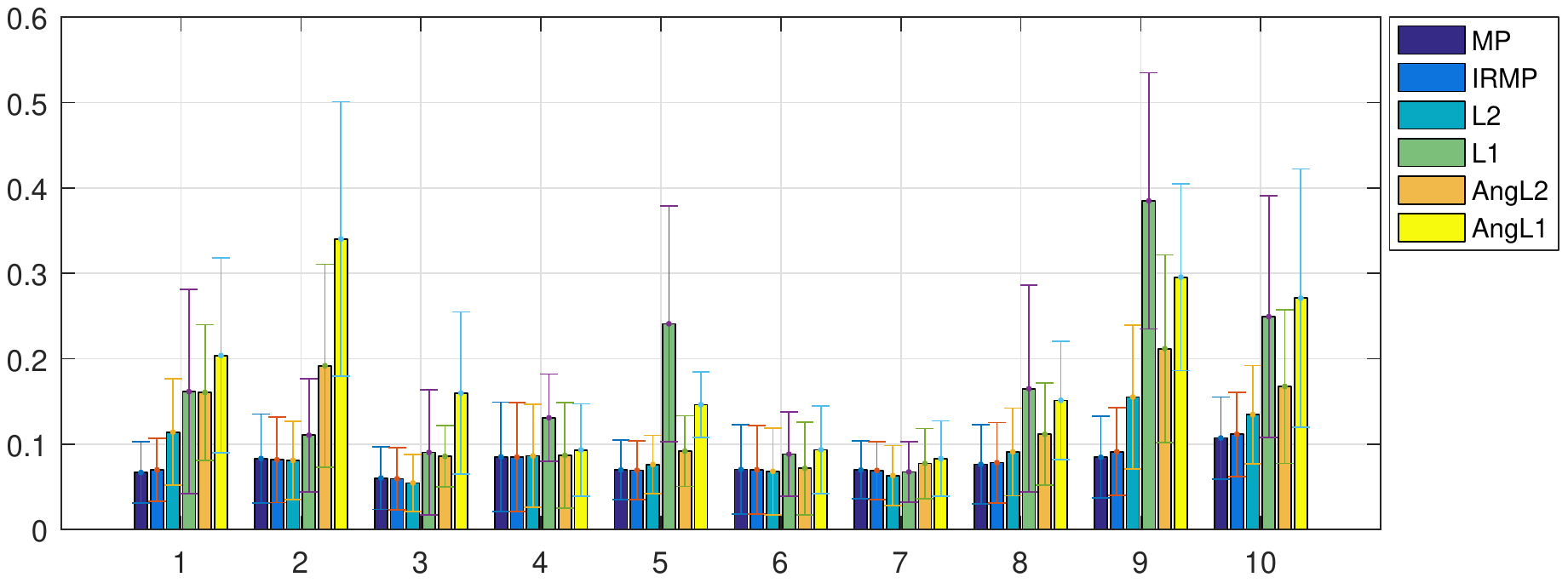}
    \caption{The mean and standard deviation of 3D triangulation errors for different methods in the first ten random runs on the synthetic dataset. The errors are computed for random 20 points selected in a region. The cameras and the region containing the points are shown in Fig.~\ref{fig:Conf4NC3}.
    %Synthetic dataset - $N_c = 3$ - 10 first runs - mean and std 3D error of 20 points
    }\label{fig:ExpF10Nc3}
%    \vspace{5mm}
\end{figure}

\begin{figure}
    \centering
    \includegraphics[width=0.45\textwidth,clip,keepaspectratio]{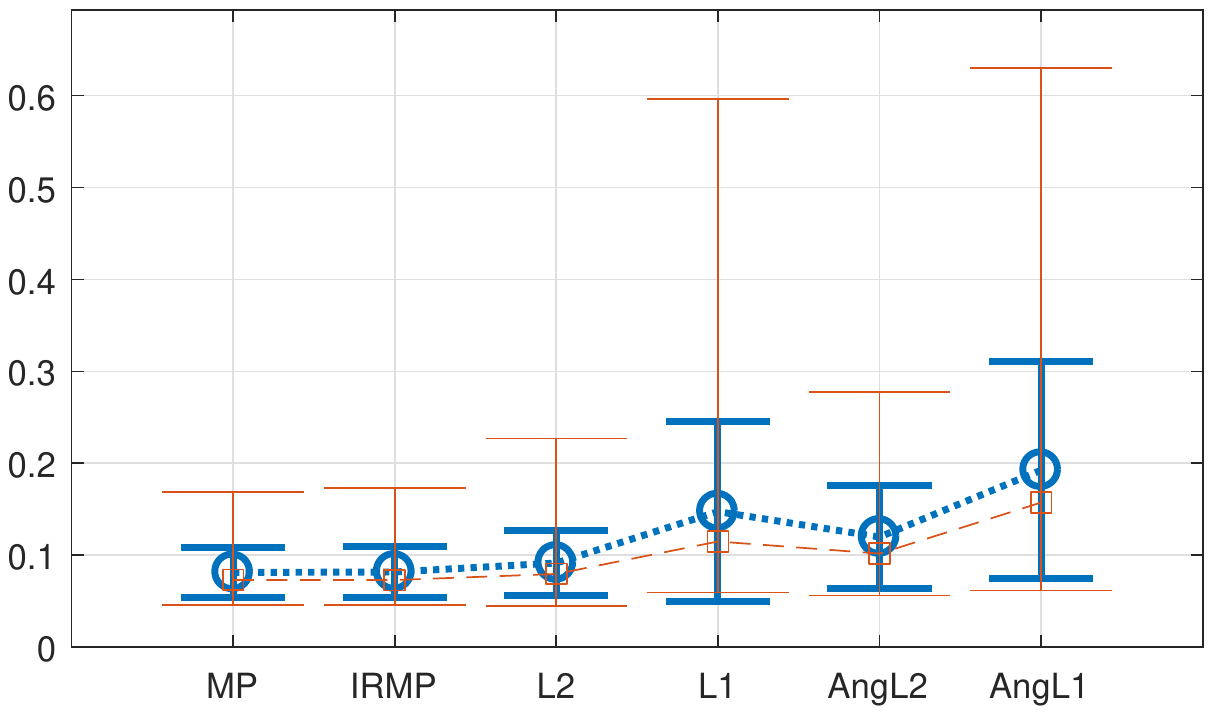}
    \caption{The mean, standard deviation, median, minimum, and maximum of 100 mean 3D triangulation error for different methods. The errors are computed for random 20 points selected in a region. The cameras and the region containing the points are shown in Fig.~\ref{fig:Conf4NC3}.
    %Synthetic dataset - $N_c = 3$ - 100 run - mean, std, median, min and max of mean 3D error
    }\label{fig:ExpF100Nc3}
\end{figure}

To evaluate the performance of different methods on multi-view triangulation, another camera is added to the mentioned two cameras setup as shown in Fig.~\ref{fig:Conf4NC3}. The mean and standard deviation of triangulation error of all $20$ points for different methods in the first~$10$ experiments are shown in Fig.~\ref{fig:ExpF10Nc3}. Again the mean, median, standard deviation, minimum, and maximum of the mean error of all $100$ experiments are computed and are shown in Fig.~\ref{fig:ExpF100Nc3}.

\subsection{Full Reconstruction Procedure in a Real Dataset}
\begin{figure}
    \centering
    \includegraphics[width=0.5\textwidth,clip,keepaspectratio]{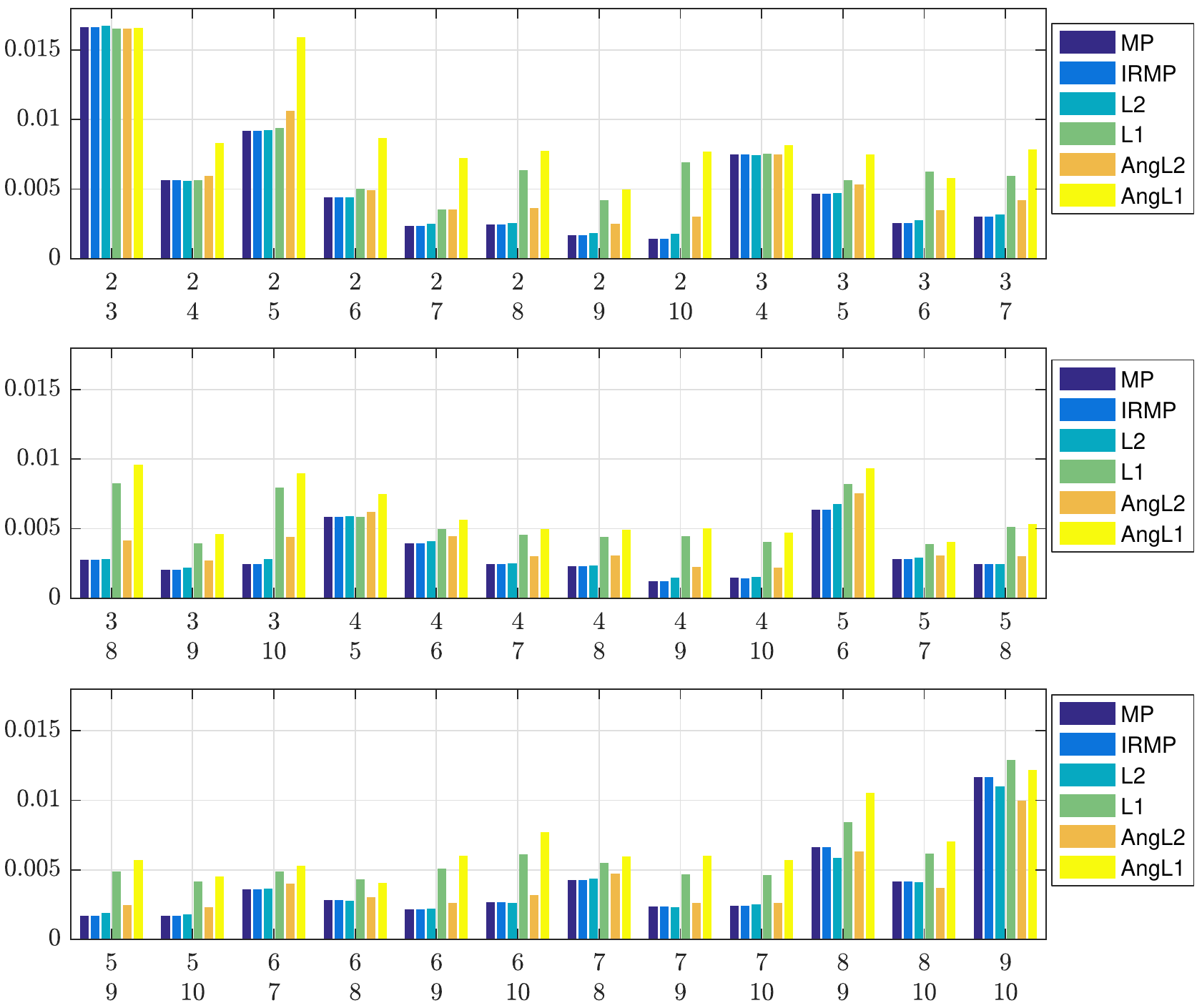}
    \caption{The mean triangulation error of different methods for each selected pairs of cameras in all 2-combination of cameras in Fountain-P11 dataset. The results are for 10 runs of the triangulation procedure, and in each run 20 random corresponding points using SURF features are used.
   % The experiments are repeated 10 times for different random 20 points.
    %The figure depicted the mean triangulation 3D error of all experiments for each selected pair cameras.
%    The comparison of different triangulation method.
%    The mean 3D triangulation error of 20 random points using different methods are calculated for each member of a set of experiments. The set of experiments consists of all 2-combinations of cameras in Fountain-P11 dataset and are repeated 10 times with different 20 random points.
%    The set of experiments consists of triangulating 20 points from all 2-combinations of cameras in Fountain-P11 dataset. and are repeated 10 times with different 20 random points.
%    For each member of all 2-combinations of cameras in Fountain-P11 dataset
%    The results are the mean, standard deviation, median, minimum, and maximum of mean 3D triangulation error in a set of experiments. The set of experiments
%    All 2-combinations of cameras in Fountain-P11 dataset are
%    The figure depicts
%    all 2 combination of cameras in Fountain-P11 dataset
%    The mean, standard deviation, median, minimum, and maximum of mean 3D triangulation error of 20 random points using different methods. The cameras and the points region are shown in Fig.~\ref{fig:Conf4NC3}.
    %Real dataset - $N_c = 2$ - 36*10 run - mean 3D error of 20 points
    }\label{fig:ExpF1RD_}
\end{figure}

\begin{figure}
    \centering
    \includegraphics[width=0.45\textwidth,clip,keepaspectratio]{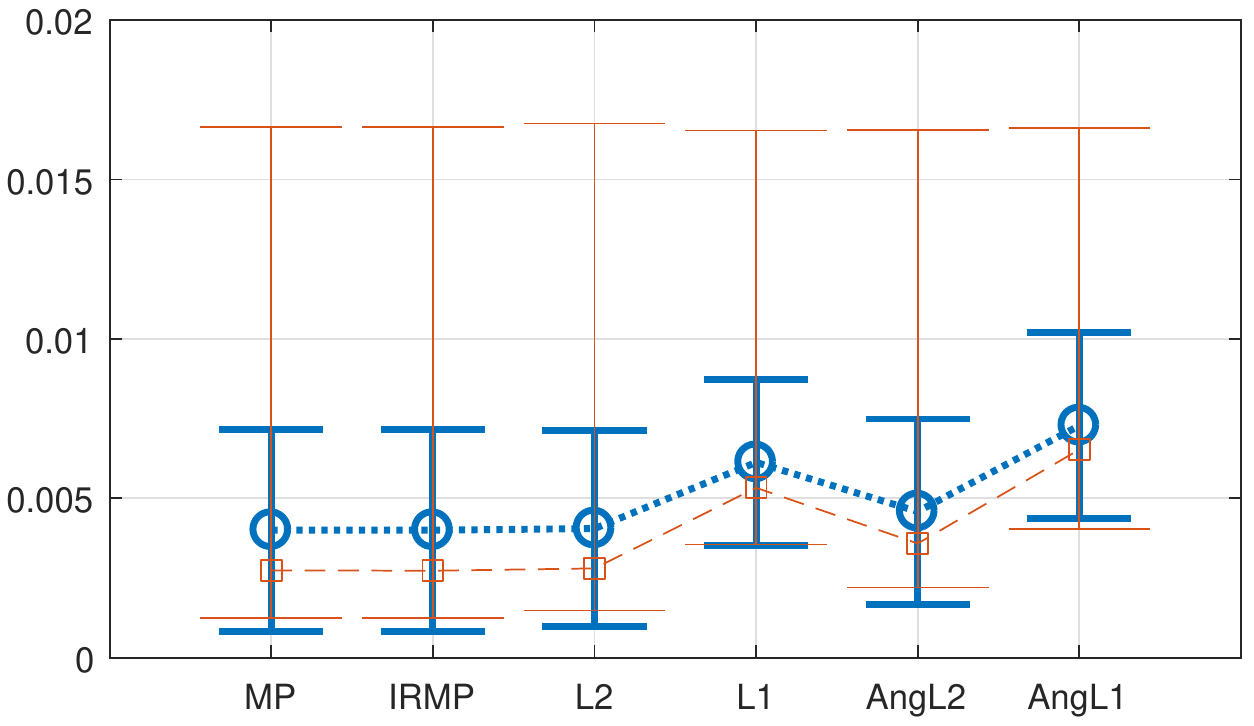}
    \caption{The mean, standard deviation, median, minimum, and maximum of mean 3D triangulation error of all 2-view experiments on Fountain-P11 dataset. %The experiments are triangulating 20 random points on all 2-combinations of cameras.
    %Real dataset - $N_c = 2$ - 36*10 run - mean, std, median, min and max of mean 3D error
    }\label{fig:ExpF1RD}
\end{figure}

In this part, ``Fountain-P11'' dataset is used to evaluate the triangulation methods. The SURF feature correspondences \cite{bay2008speeded} are used to find the essential matrices between all $2$-combinations of cameras\footnote{Fountain-P11 dataset consists of $11$ images from different perspectives. We remove the first and last images, which have a few number of feature correspondences and find the essential matrices for all $2$-combinations of all other $9$ images.}. The test process is the same as the process in the synthetic datasets except that the projection in step~$2$ is replaced by the feature matching and no noise is added to the points anymore. For each selected pair of cameras, the process is repeated $10$ times for different random corresponding points. Figs.~\ref{fig:ExpF1RD_} and \ref{fig:ExpF1RD} show the results of different triangulation methods in the mentioned process.

\begin{figure}[t]
    \centering
    \includegraphics[width=0.45\textwidth,clip,keepaspectratio]{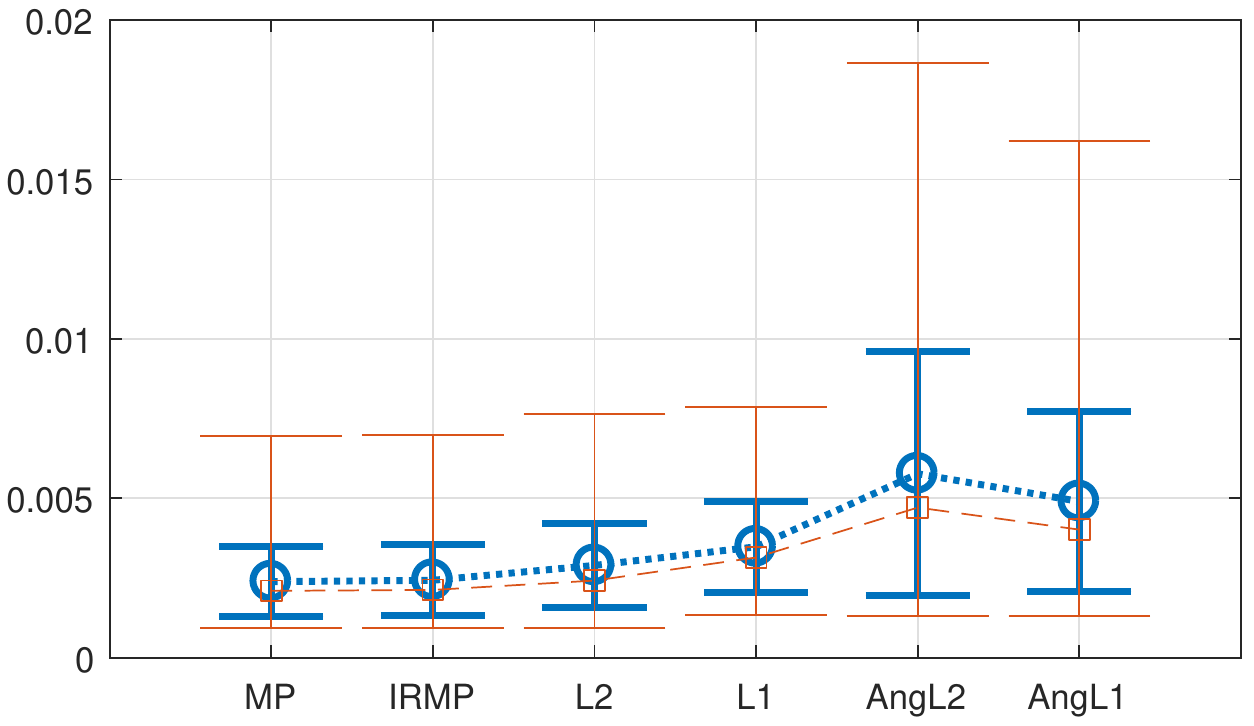}
    \caption{The mean, standard deviation, median, minimum, and maximum of mean 3D triangulation error of all 3-view experiments on Fountain-P11 dataset.
    %Real dataset - $N_c = 3$ - 84*10 run - mean, std, median, min and max of mean 3D error
    }\label{fig:ExpF1RDNc3}
\end{figure}

The experiment is repeated for triangulation by three cameras. Again the process is the same as the process in the synthetic datasets of subsection~\ref{More Than Two Cameras} with feature matching used in step 2. The process is repeated $10$ time for any $3$-combinations of cameras. The results are shown in Fig.~\ref{fig:ExpF1RDNc3}.
As it can be seen in the results, the mid-point method outperforms the other methods and has less mean 3D reconstruction error in all the experiments.


\section{Conclusion}
In this paper, different triangulation methods were evaluated in terms of 3D reconstruction accuracy in a calibrated \textit{structure-from-motion} setting. It was shown that the mid-point triangulation method, which has a closed-form solution for any number of cameras, is less sensitive to error in the cameras extrinsic parameters in comparison to the other methods. This results in a better performance of this triangulation method in \textit{structure-from-motion} procedures. The performance of different methods in a \textit{structure-from-motion} process were evaluated in synthetic and real datasets through extensive experiments. It was shown that the mid-point triangulation method outperforms the commonly used $L_2$ triangulation method \cite{hartley1997triangulation} in typical practical applications, where cameras extrinsic parameters are computed based on image registration and consequently have uncertainties.

\bibliographystyle{spmpsci}
\bibliography{refs}

\begin{thebibliography}{10}
\providecommand{\url}[1]{{#1}}
\providecommand{\urlprefix}{URL }
\expandafter\ifx\csname urlstyle\endcsname\relax
  \providecommand{\doi}[1]{DOI~\discretionary{}{}{}#1}\else
  \providecommand{\doi}{DOI~\discretionary{}{}{}\begingroup
  \urlstyle{rm}\Url}\fi

\bibitem{arrigoni2018robust}
Arrigoni, F., Rossi, B., Fragneto, P., Fusiello, A.: Robust synchronization in
  {SO(3)} and {SE(3)} via low-rank and sparse matrix decomposition.
\newblock Computer Vision and Image Understanding \textbf{174}, 95--113 (2018)

\bibitem{bartoli2005structure}
Bartoli, A., Sturm, P.: Structure-from-motion using lines: Representation,
  triangulation, and bundle adjustment.
\newblock Computer Vision and Image Understanding \textbf{100}(3), 416--441
  (2005)

\bibitem{bay2008speeded}
Bay, H., Ess, A., Tuytelaars, T., Van~Gool, L.: Speeded-up robust features
  {(SURF)}.
\newblock Computer Vision and Image Understanding \textbf{110}(3), 346--359
  (2008)

\bibitem{byrod2007improving}
Byr{\"o}d, M., Josephson, K., {\AA}str{\"o}m, K.: Improving numerical accuracy
  of {G}r{\"o}bner basis polynomial equation solvers.
\newblock In: IEEE International Conference on Computer Vision, pp. 449--456.
  IEEE (2007)

\bibitem{castle2011wide}
Castle, R.O., Klein, G., Murray, D.W.: Wide-area augmented reality using camera
  tracking and mapping in multiple regions.
\newblock Computer Vision and Image Understanding \textbf{115}(6), 854--867
  (2011)

\bibitem{chatterjee2017robust}
Chatterjee, A., Govindu, V.M.: Robust relative rotation averaging.
\newblock IEEE Transactions on Pattern Analysis and Machine Intelligence
  \textbf{40}(4), 958--972 (2017)

\bibitem{hartley2007optimal}
Hartley, R., Kahl, F.: Optimal algorithms in multiview geometry.
\newblock In: Asian Conference on Computer Vision, pp. 13--34. Springer (2007)

\bibitem{hartley2013verifying}
Hartley, R., Kahl, F., Olsson, C., Seo, Y.: Verifying global minima for {L$_2$}
  minimization problems in multiple view geometry.
\newblock International Journal of Computer Vision \textbf{101}(2), 288--304
  (2013)

\bibitem{hartley2004sub}
Hartley, R., Schaffalitzky, F.: L$_\infty$ minimization in geometric
  reconstruction problems.
\newblock In: IEEE Conference on Computer Vision and Pattern Recognition,
  vol.~1, pp. 504--509. IEEE (2004)

\bibitem{hartley2013rotation}
Hartley, R., Trumpf, J., Dai, Y., Li, H.: Rotation averaging.
\newblock International Journal of Computer Vision \textbf{103}(3), 267--305
  (2013)

\bibitem{hartley1997triangulation}
Hartley, R.I., Sturm, P.: Triangulation.
\newblock Computer Vision and Image Understanding \textbf{68}(2), 146--157
  (1997)

\bibitem{jiang2013global}
Jiang, N., Cui, Z., Tan, P.: A global linear method for camera pose
  registration.
\newblock In: IEEE International Conference on Computer Vision, pp. 481--488
  (2013)

\bibitem{kanatani1996statistical}
Kanatani, K.: Statistical optimization for geometric computation: theory and
  practice.
\newblock Elsevier, New York, USA (1996)

\bibitem{kukelova2008polynomial}
Kukelova, Z., Bujnak, M., Pajdla, T.: Polynomial eigenvalue solutions to the
  5-pt and 6-pt relative pose problems.
\newblock In: British Machine Vision Conference, vol.~2, pp. 56.1--56.10 (2008)

\bibitem{lee2019closed}
Lee, S.H., Civera, J.: Closed-form optimal two-view triangulation based on
  angular errors.
\newblock In: IEEE International Conference on Computer Vision, pp. 2681--2689
  (2019)

\bibitem{li2007practical}
Li, H.: A practical algorithm for {L$_\infty$} triangulation with outliers.
\newblock In: IEEE Conference on Computer Vision and Pattern Recognition, pp.
  1--8. IEEE (2007)

\bibitem{lindstrom2010triangulation}
Lindstrom, P.: Triangulation made easy.
\newblock In: IEEE Conference on Computer Vision and Pattern Recognition, pp.
  1554--1561. IEEE (2010)

\bibitem{lo2009local}
Lo, T.W.R., Siebert, J.P.: Local feature extraction and matching on range
  images: {2.5D SIFT}.
\newblock Computer Vision and Image Understanding \textbf{113}(12), 1235--1250
  (2009)

\bibitem{lowe1999object}
Lowe, D.G.: Object recognition from local scale-invariant features.
\newblock In: IEEE International Conference on Computer Vision, vol.~2, pp.
  1150--1157. IEEE (1999)

\bibitem{nister2004efficient}
Nister, D.: An efficient solution to the five-point relative pose problem.
\newblock IEEE Transactions on Pattern Analysis and Machine Intelligence
  \textbf{26}(6), 756--770 (2004)

\bibitem{olsson2010outlier}
Olsson, C., Eriksson, A., Hartley, R.: Outlier removal using duality.
\newblock In: IEEE Conference on Computer Vision and Pattern Recognition, pp.
  1450--1457. IEEE (2010)

\bibitem{ozyesil2015robust}
Ozyesil, O., Singer, A.: Robust camera location estimation by convex
  programming.
\newblock In: IEEE Conference on Computer Vision and Pattern Recognition, pp.
  2674--2683 (2015)

\bibitem{ramalingam2006generic}
Ramalingam, S., Lodha, S.K., Sturm, P.: A generic structure-from-motion
  framework.
\newblock Computer Vision and Image Understanding \textbf{103}(3), 218--228
  (2006)

\bibitem{schonberger2016structure}
Schonberger, J.L., Frahm, J.M.: Structure-from-motion revisited.
\newblock In: IEEE Conference on Computer Vision and Pattern Recognition, pp.
  4104--4113 (2016)

\bibitem{sim2006removing}
Sim, K., Hartley, R.: Removing outliers using the {L$_\infty$} norm.
\newblock In: IEEE Conference on Computer Vision and Pattern Recognition,
  vol.~1, pp. 485--494. IEEE (2006)

\bibitem{stewenius2005hard}
Stewenius, H., Schaffalitzky, F., Nister, D.: How hard is {3-view}
  triangulation really?
\newblock In: IEEE International Conference on Computer Vision, vol.~1, pp.
  686--693 (2005)

\bibitem{sweeney2015optimizing}
Sweeney, C., Sattler, T., Hollerer, T., Turk, M., Pollefeys, M.: Optimizing the
  viewing graph for structure-from-motion.
\newblock In: IEEE International Conference on Computer Vision, pp. 801--809
  (2015)

\bibitem{tippetts2016review}
Tippetts, B., Lee, D.J., Lillywhite, K., Archibald, J.: Review of stereo vision
  algorithms and their suitability for resource-limited systems.
\newblock Journal of Real-Time Image Processing \textbf{11}(1), 5--25 (2016)

\bibitem{toldo2015hierarchical}
Toldo, R., Gherardi, R., Farenzena, M., Fusiello, A.: Hierarchical
  structure-and-motion recovery from uncalibrated images.
\newblock Computer Vision and Image Understanding \textbf{140}, 127--143 (2015)

\bibitem{torr2000mlesac}
Torr, P.H., Zisserman, A.: Mlesac: A new robust estimator with application to
  estimating image geometry.
\newblock Computer Vision and Image Understanding \textbf{78}(1), 138--156
  (2000)

\bibitem{yang2019iteratively}
Yang, K., Fang, W., Zhao, Y., Deng, N.: Iteratively reweighted midpoint method
  for fast multiple view triangulation.
\newblock IEEE Robotics and Automation Letters \textbf{4}(2), 708--715 (2019)

\bibitem{zhu2018very}
Zhu, S., Zhang, R., Zhou, L., Shen, T., Fang, T., Tan, P., Quan, L.: Very
  large-scale global sfm by distributed motion averaging.
\newblock In: IEEE Conference on Computer Vision and Pattern Recognition, pp.
  4568--4577 (2018)

\end{thebibliography}
% Non-BibTeX users please use
%\begin{thebibliography}{}
%%
%% and use \bibitem to create references. Consult the Instructions
%% for authors for reference list style.
%%
%\bibitem{RefJ}
%% Format for Journal Reference
%Author, Article title, Journal, Volume, page numbers (year)
%% Format for books
%\bibitem{RefB}
%Author, Book title, page numbers. Publisher, place (year)
%% etc
%\end{thebibliography}

\end{document}